\documentclass[conference]{IEEEtran}
\IEEEoverridecommandlockouts
\usepackage{cite}
\usepackage{amsmath,amssymb,amsfonts}
\usepackage{algorithmic}
\usepackage{graphicx}
\usepackage{textcomp}
\usepackage{xcolor}
\usepackage{bm}
\usepackage[font={small}]{caption}
\usepackage{subcaption}
\usepackage{hyperref}

\def\BibTeX{{\rm B\kern-.05em{\sc i\kern-.025em b}\kern-.08em
    T\kern-.1667em\lower.7ex\hbox{E}\kern-.125emX}}
\begin{document}

\title{An Autonomous Probing System \\for Collecting Measurements at Depth \\from Small Surface Vehicles}

\author{\parbox{\linewidth}{\centering
 Yuying Huang$^{*,1}$,
Yiming Yao$^{*,1}$, 
Johanna Hansen$^{*,1}$, \\
Jeremy Mallette$^{1}$, 
Sandeep Manjanna$^{1,2}$, 
Gregory Dudek$^{1}$, and
David Meger$^{1}$\\ \vspace{.4cm}
 $^{1}$\textit{Mobile Robotics Lab, McGill University}, Montreal, Canada \\
$^{2}$\textit{Computer and Information Science, University of Pennsylvania}, Philadelphia, USA
\thanks{$^{*}$indicates equal contribution}
}}   
\maketitle
\begin{abstract}
This paper presents the portable autonomous probing system (APS), a low-cost robotic design for collecting water quality measurements at targeted depths from an autonomous surface vehicle (ASV). This system fills an important, but often overlooked niche in marine sampling by enabling mobile sensor observations throughout the near-surface water column without the need for advanced underwater equipment.
We present a probe delivery mechanism built with commercially available components and describe the corresponding open-source simulator and winch controller.   Finally, we demonstrate the system in a field deployment and discuss design trade-offs and areas for future improvement. 
Project details are available on \href{https://johannah.github.io/publication/sample-at-depth/}{\color{blue} our website \\}
\end{abstract}


\section{Introduction}
Active monitoring of water resources such as estuaries, lakes, and bays is of importance for improving our understanding of how water quality impacts natural ecosystems and human populations. Our knowledge of these systems is often limited by the expense of gathering frequent and diverse samples. While there has been wide adoption of autonomous boats\cite{survey_ASV} for capturing in situ data from the surface, underwater observations are mostly out of reach.    

Our work is motivated by the desire to increase the availability of vertical profiles in the water column in small bodies of water. The need for in situ data from throughout the water column has been well-established in the oceanographic \cite{doherty1999moored} and limnological  \cite{TurbidityPhytoplankton} communities. The demand for this water-column data is perhaps best illustrated by the incredible success and adoption of data from the ARGO profiling float program~\cite{ARGO_profiler}. In lakes, the need for portable depth profiling systems is clearly motivated. Lofton \emph{et al.}~\cite{Beatrix} employ water-column lake observations of nutrients, temperature, and light to investigate phytoplankton distribution in over $51$ northern lakes.

\begin{figure}[h]
\begin{subfigure}{.6\columnwidth}
\centering
  \includegraphics[width=2in,height=1.5in]{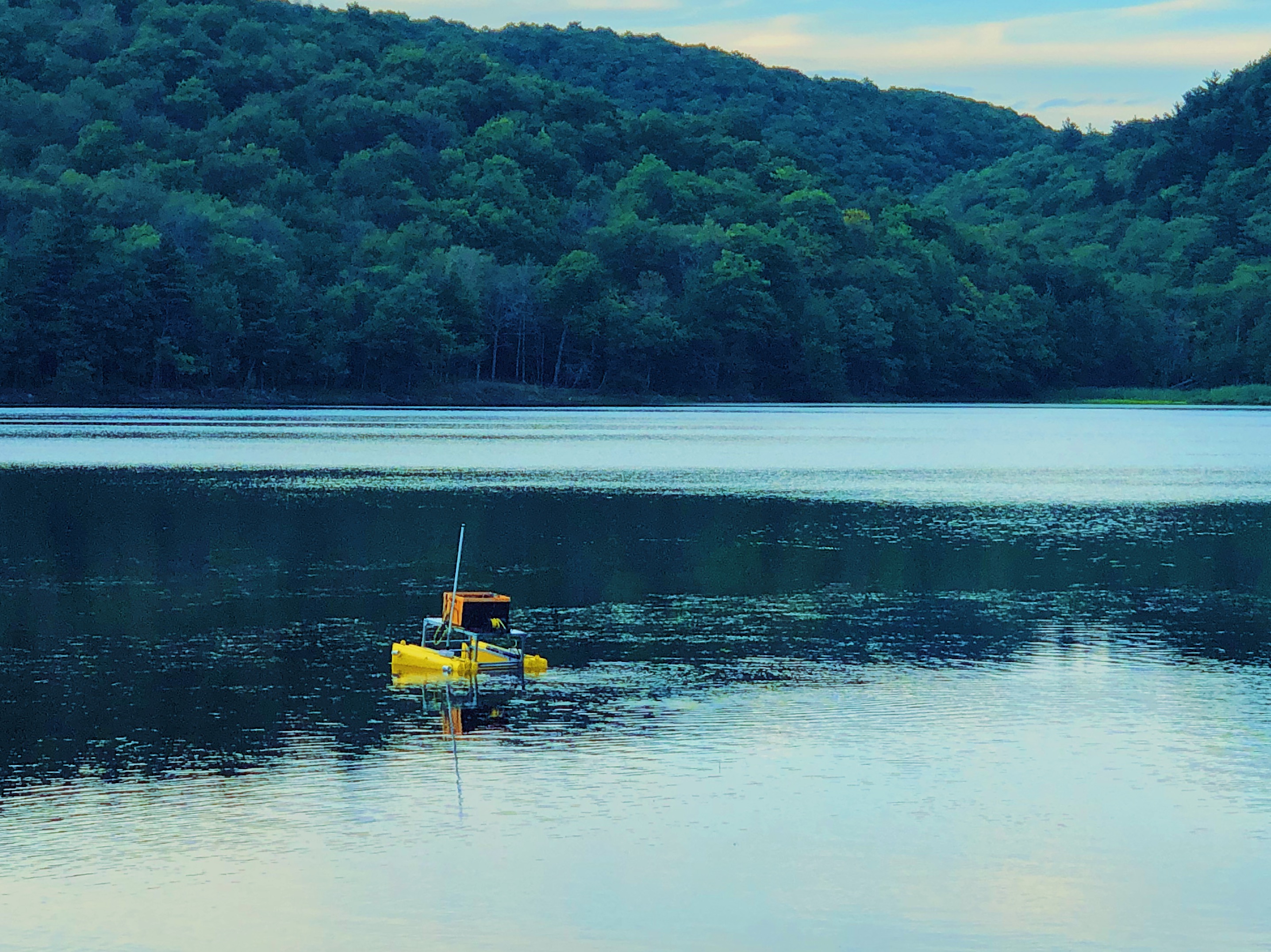}
  \caption{ASV}
  \label{fig:beauty}
\end{subfigure}
\begin{subfigure}{.3\columnwidth}
\centering
  \includegraphics[width=1in,height=1.5in]{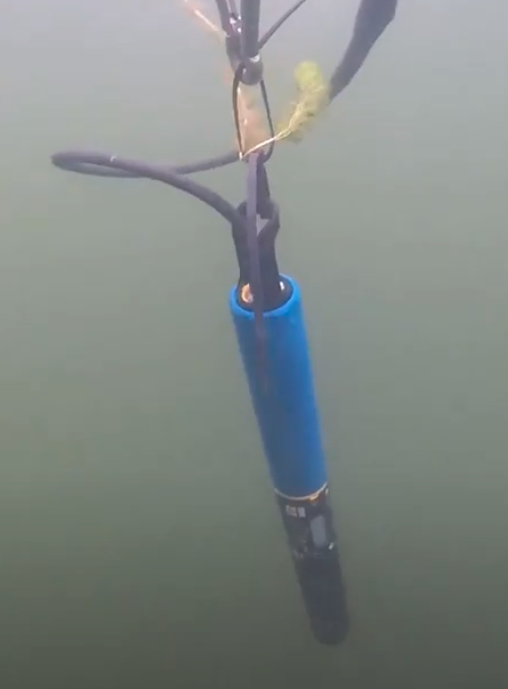}
  \caption{Probe}
  \label{fig:beauty}
  \end{subfigure}
  \caption{This figure captures our autonomous probing system (APS) deploying a water quality sensor called a sonde to a specified depth in the water column. }
\end{figure}

\begin{figure*}[h]
\centering
\begin{subfigure}{.4\textwidth}
  \centering
  \includegraphics[width=\linewidth,height=1.5in]{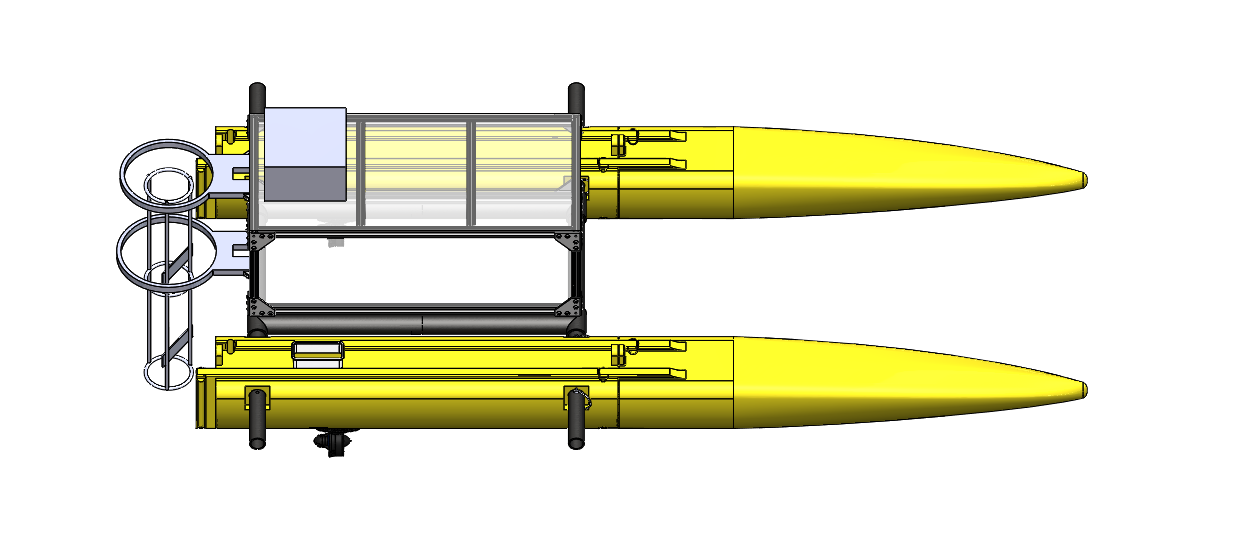}
  \caption{CAD Diagram}
  \label{fig:CAD}
\end{subfigure}
\begin{subfigure}{.2\textwidth}
  \centering
  \includegraphics[width=0.85\linewidth,height=1.5in]{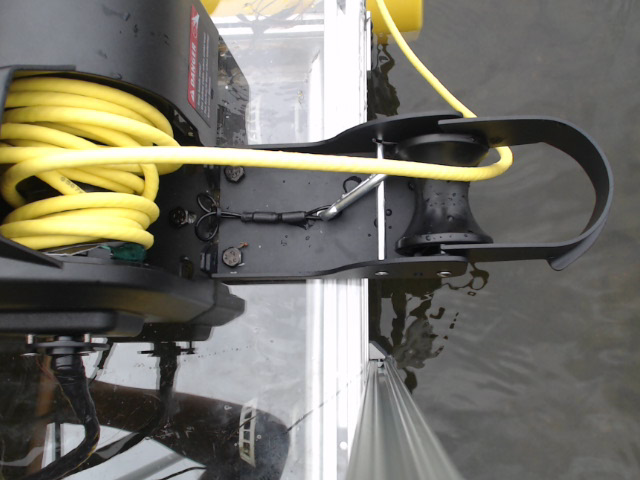}
  \caption{Winch}
  \label{fig:winchcam}
\end{subfigure}
\hfill
\begin{subfigure}{.17\textwidth}
  \centering
  \includegraphics[,height=1.5in, width=\linewidth, trim=600 200 500 300,clip]{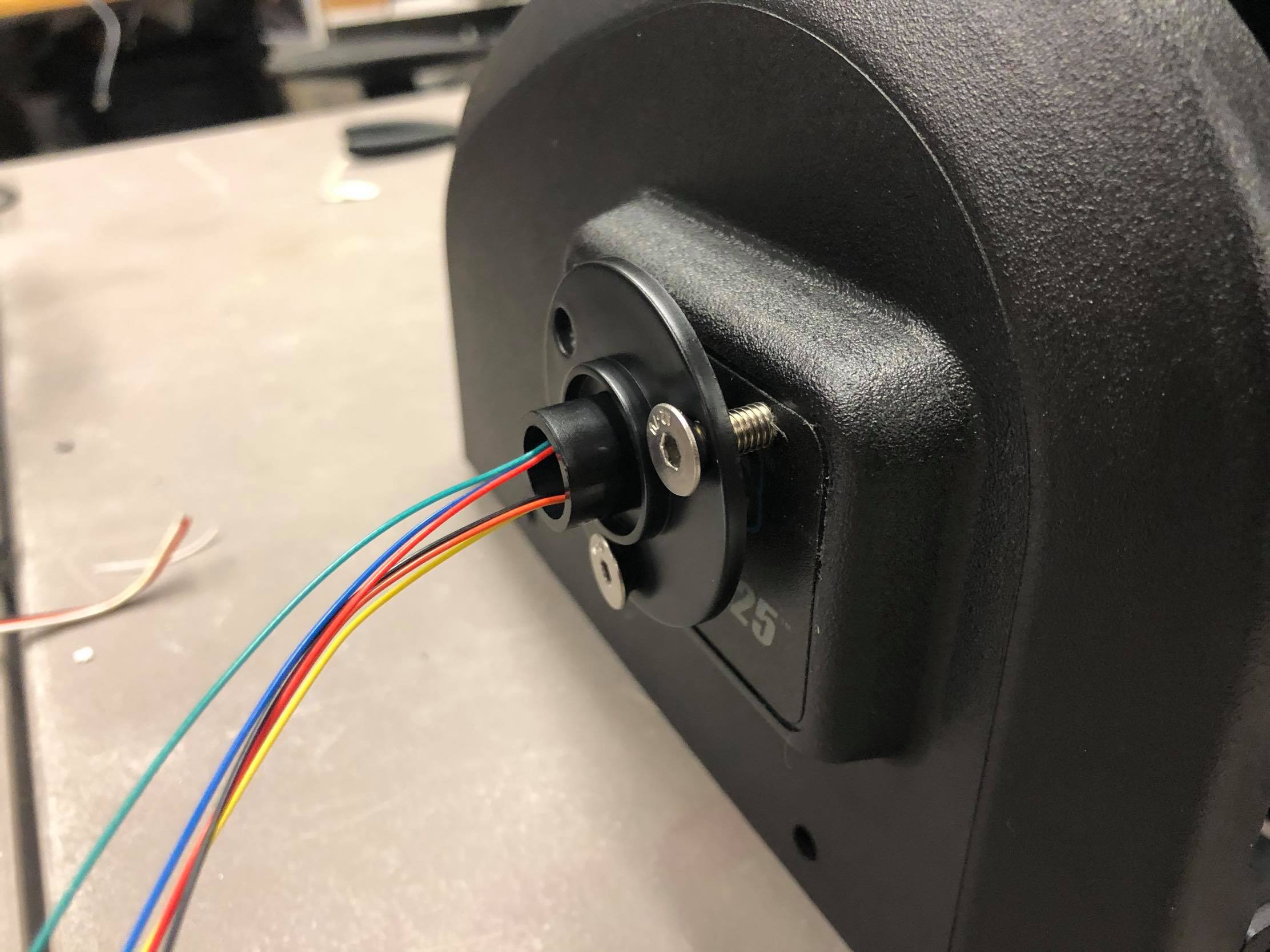}
  \caption{Slip Ring}
  \label{fig:slip}
\end{subfigure}
\hfill
\begin{subfigure}{.17\textwidth}
  \centering
  \includegraphics[width=\linewidth,height=1.5in]{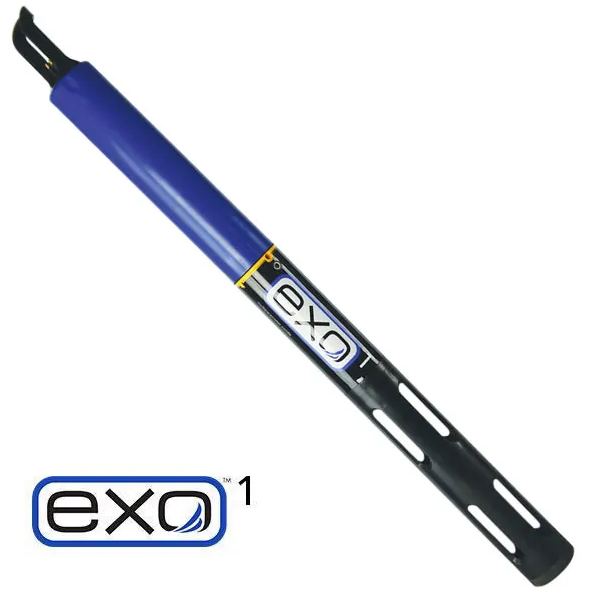}
  \caption{EXO Sensor}
  \label{fig:sonde}
\end{subfigure}
\caption{Fig.~\ref{fig:CAD} shows the CAD diagram of the entire system we used in our experiments and in our FEA analysis.  Fig.~\ref{fig:winchcam} shows the commercially available winch that we utilized for probe deployment and retrieval. This view of the winch and tether is from the \emph{winchcam}, a camera mounted directly above the winch to provide remote monitoring. We add an electronic slip ring (Fig.~\ref{fig:slip}) to the winch and an underwater tether to enable continuous communication with the probe. In our experiments, we employ the EXO Sonde as our probe, depicted in Fig.~\ref{fig:sonde}.}
\label{fig:hardware}
\end{figure*}

This paper presents the design and demonstration of a low-cost system to capture samples to valuable near-surface water column data (up to $20~m$).  Our platform, consisting of a modular frame, off-the-shelf winch, and sensor probe (depicted in Fig.~\ref{fig:hardware}) mounts on a fully controllable autonomous surface vehicle (ASV). In the following sections, we describe the mechanical design (Section \ref{sec:design}),  present a new open-source simulator (Section \ref{sec:simulator}), and finally, we present results from field trials (Section \ref{sec:experiments}) and discuss design limitations and room for improvement (Section \ref{sec:limitation}).

There have been consistent efforts to reduce the cost (both human and monetary) of acquiring marine data. Shallow ($0-20~m$) marine phenomena are traditionally measured either by passive sensors, such as fixed moorings  or drifting nodes such as the aforementioned profiling floats \cite{mooringsanddrifters, ARGO_profiler}. While these are powerful sources of data at depth, both are limited in where they can sample. Moorings are typically deployed to a static location and can be expensive to relocate. Drifting sensors cannot be explicitly utilized for actively gathering information from scientifically important positions as they are not spatially controllable. 

Underwater vehicles, such as \cite{shallow_auv, fish} enable highly specific sampling throughout the water column, but are limited by their sophisticated nature. Expert engineers are typically needed to deploy these costly systems, which may not have access to satellite positioning. Due to the lack of accessible, highly maneuverable, and cost-effective methods for sampling from the water column across a body of water, scientists often fall back to laborious manual data acquisition campaigns \cite{methods}.

\section{Background} 

Unmanned surface vehicles have found increasing prominence in near-shore surface observation operations~\cite{hydronet, MadeoASV, Pradalier2019}. These inexpensive and intuitive systems fill various niches such as persistent monitoring \cite{NUSwan}, collecting in situ collections of both physical samples \cite{sandeep_water, gene_sample, adaptive_ecosystem} and  observational samples \cite{Saildrone, sandeep_policy}. 

Sensors deployed from ships have a long history  of extending observational capabilities from manned research vessels by deploying tethered sensors \cite{towed-titanic}. ASVs have also been utilized to manage tethered underwater vehicles \cite{jung2018study}, effectively serving as a local localization and communication station for advanced vehicles.  

We are interested in developing a budget-friendly resource for scientists to autonomously capture data from below the surface. This makes the ASV a natural platform from which to transport and deliver our sensing platform. Others have employed a similar approach. In \cite{hitz2012design}, the authors develop an advanced probe which is capable of controlling its position within the water column. The design in \cite{ASV_depth} presents a winch and probe system for collecting physical water samples along the water column at up to $50~m$ depth on the \cite{hydronet} platform. 

Our system, inspired by both \cite{ASV_depth} and \cite{hitz2012design}, is a simpler and modern alternative which utilizes commercially available components to deliver a probe to controllable near-surface depths. We 
envision that this flexible and approachable mechanical design will increase accessibility for near-shore science missions.


\begin{figure}[h]
    \centering
    \includegraphics[width=.8\linewidth]{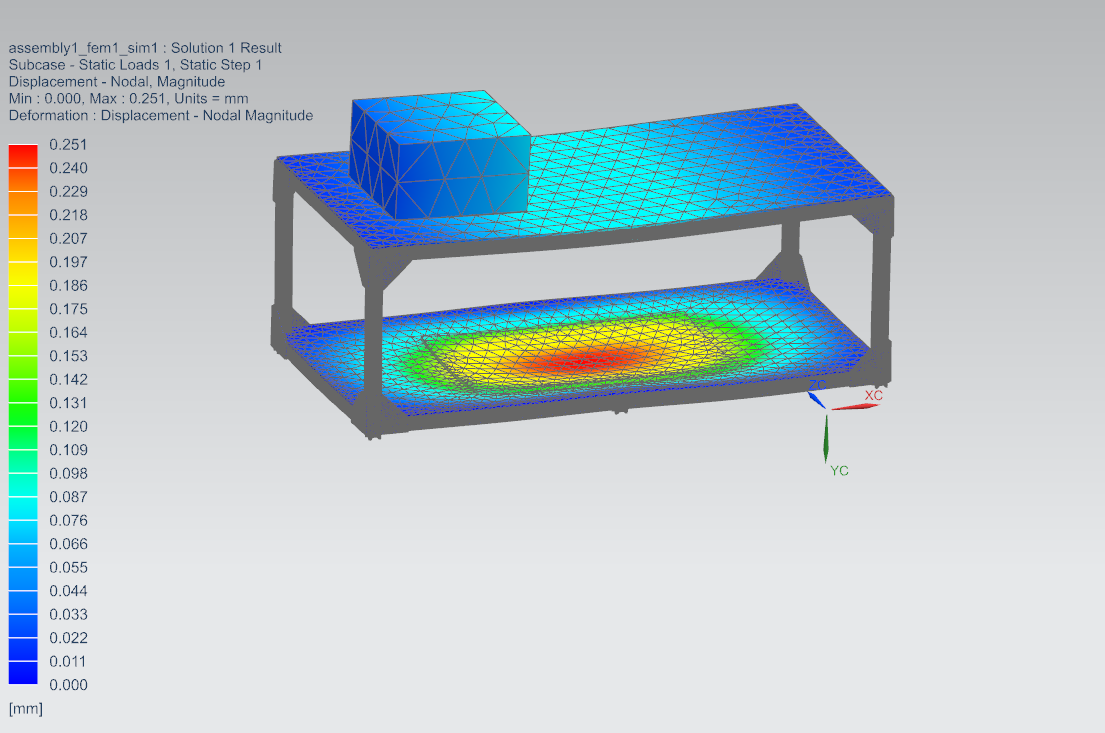}
    \caption{Displacement map from Finite Element Analysis, showing highest concentrations at the centre of each plate}
    \label{fig:FEA1}
\end{figure}
\begin{figure*}[h]
\begin{subfigure}{.52\linewidth}
    \centering
    \includegraphics[width=3.2in,height=1.5in]{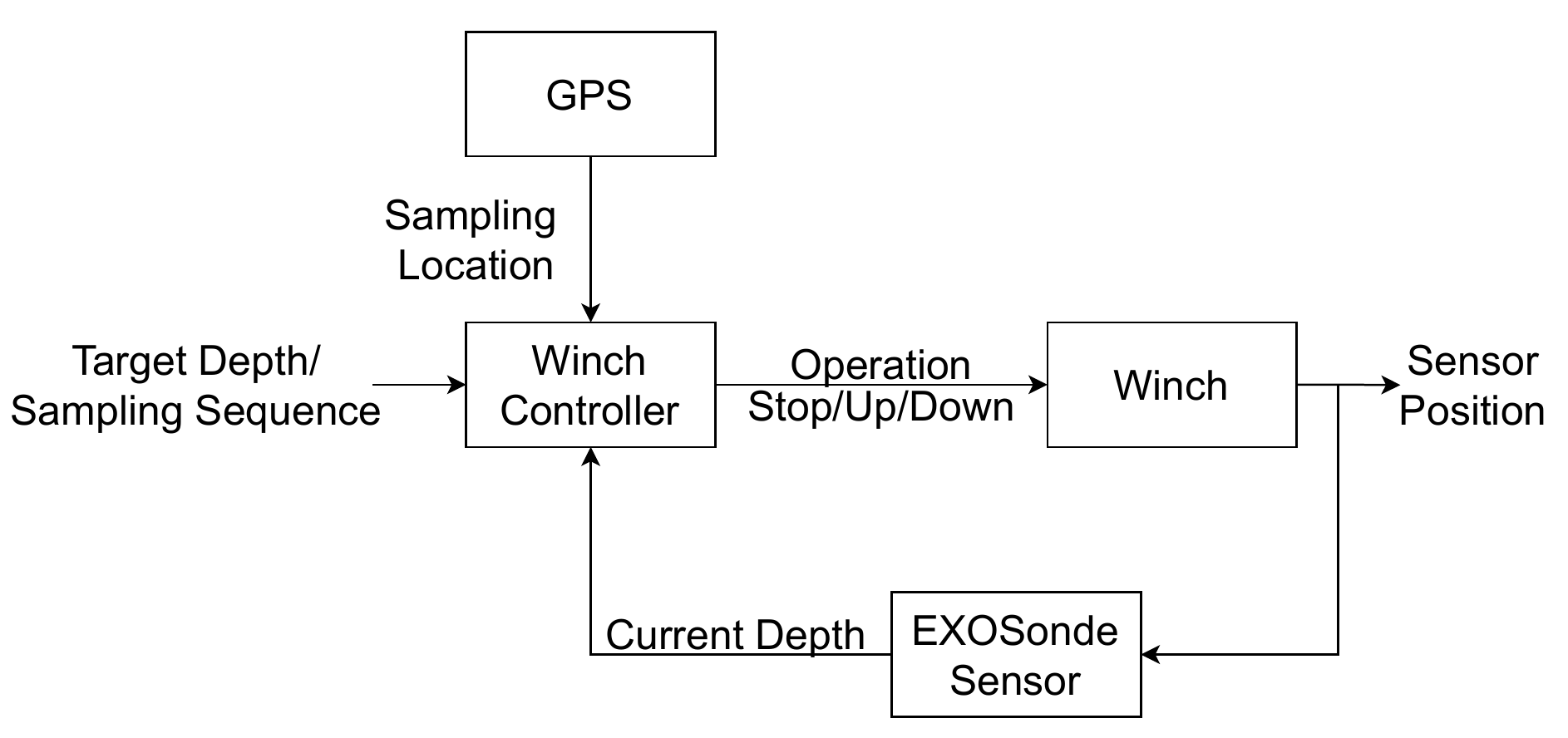}
    \caption{Controller diagram}
    \label{fig:controller}
    \end{subfigure}
\begin{subfigure}{.3\linewidth}
    \centering
    \includegraphics[width=1.9in,height=1.5in]{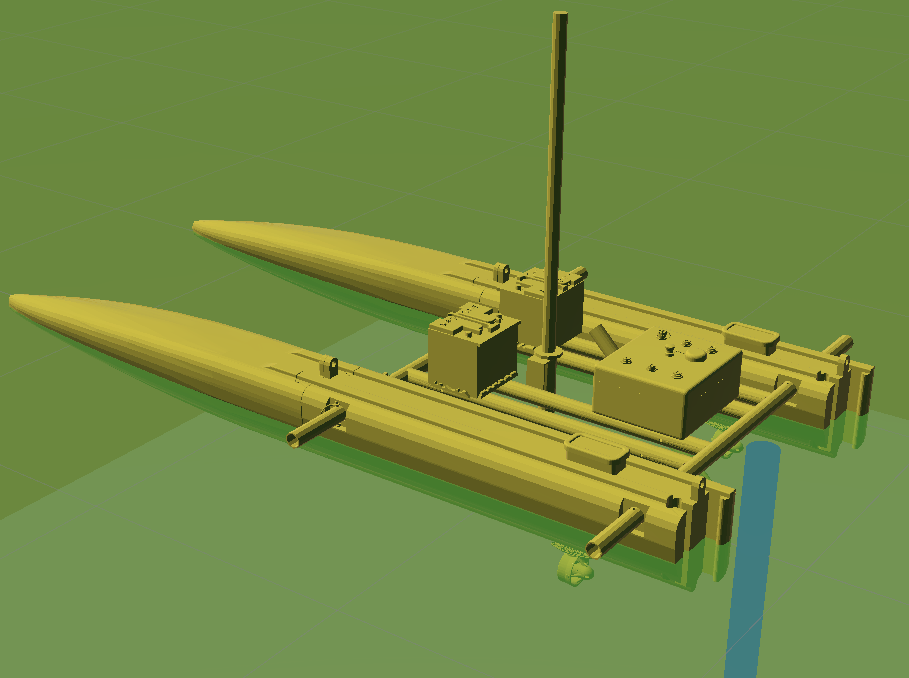}
    \caption{Unity simulation}
    \label{fig:simulator}
\end{subfigure}
\begin{subfigure}{.13\linewidth}
    \centering
    \includegraphics[width=.9in,height=1.5in]{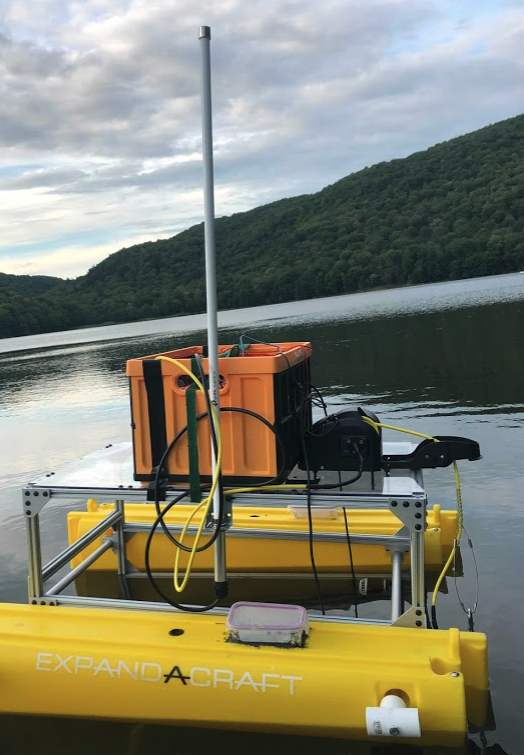}
    \caption{Configured ASV}
    \label{fig:conf-asv}
\end{subfigure}
\caption{Fig.~\ref{fig:controller} details the control system which governs the depth control for our winch. A fully-controllable version of our system is available in our simulator, shown in Fig.~\ref{fig:simulator}. Finally, \ref{fig:conf-asv} shows our fully-configured system deployed.}
\end{figure*}

\section{System Design}
\label{sec:design}
This autonomous probing system was designed with the goal of demonstrating an approachable configuration for collecting in situ samples at depth from surface vehicles. Although we present field results on a small catamaran-style ASV, our modular design is compatible with many platforms. 

We approached the problem of designing the probe deployment system with the following objectives: 

\begin{itemize}
    \item \textbf{Size:} small and lightweight to enable transport by $2-3$ persons and transportation by commercial airline.
    \item \textbf{Compatibility:} the probe is interchangeable.  
    \item \textbf{Safety:} the winch controller should prioritize safety of the entire system and fail in a non-catastrophic way. 
    \item \textbf{Highly Configurable:} the frame and mounting system should be adaptable as new equipment is added or old equipment is removed. This controller can be easily integrated with any ROS-based system. 
    \item \textbf{Reproducible and Open} The system is developed from commercially available parts and the winch control software is available open-source. 
\end{itemize}

The system is sized to be installed on a small (human-portable) boat. The design was built to be highly maneuverable and re-configurable so that it can be transported both by air and in personal vehicles. The main components, consisting of a frame, winch, and probe are detailed in Fig. \ref{fig:hardware}.

\subsection{Frame}
The mounting frame is constructed from the highly-configurable and anti-corrosive $80-20$ T-slot aluminum profiles. In our experiments, the $80-20$ frame is mounted on two $0.0642$ $m^{3}$ Polyethylene pontoons. In this configuration, the flotation is limited to $106.7~kg$ in the fresh water (assumed density of 997 $kg/m^{3}$) with a safety factor of $1.2$ based on the buoyancy force as described by Eq.~\ref{eq:buoyant}.  
\newpage
\begin{equation}\label{eq:buoyant}
    F_{b} =\rho_{water}\times g \times V_{pvc} = 1256.8 N
\end{equation}

\begin{equation}
    W_{overall} = \dfrac{F_{b}}{S.F}\div g = 106.76 kg
\end{equation}

There are several key advantages in the design of the mounting frame. The horizontal T-slot profiles attach to the vertical T-slots, meaning the number and height of frame layers is flexible. This design choice was imperative, as we find it is often necessary to adjust the ASV's configuration as new instruments are included in the system.  With the $80/20$ design, we can easily add additional $80/20$-compatible T-slots, fasteners, or T-studs when new configurations are needed.  
The frame attaches to two aluminum rods which assemble into the $1.5~in$ native holes in our pontoons (which are actually commercially-available kayak stability floats). To secure the rods to the floats, we reconfigure $1.5~in$  T-shaped female PVC into \emph{endcaps} by drilling  $8mm$ bolts into the T intersection.  To attach the endcaps we  press-fit $8mm$ nuts into the inside of the aluminum rods. This ergonomic endcap  makes the boat easy to assemble, sturdy, and enables fast replacement of a pontoon in the water if necessary. 



Finite Element Analysis (FEA) was performed in Siemens’ NX software to ensure the overall structure can withstand the applied loads. Our winch located at the back upper layer is modeled as a box with the dimension of $~40.6~\times~30.5~\times~20.3~cm$ and weights $5.9~kg$ according to the user manual. Two golf-cart batteries that weigh $50~kg$ in total were placed on the bottom plate. The two $1.2192~m$ long aluminum supporting rods are assumed rigid. The force exerted on the winch by the tether and the probe and gravitational force acting on all parts were taken into account in this analysis.

The simulated stresses on the frame, according to our described load, reaches a maximum stress well below the ultimate strength of the material ($120~MPa$ for $6051$ aluminum); this indicates high reliability ($8.16$ safety factor calculated via Eq.~\ref{eq:safet}) and the option to add additional payload. We found that the maximum stress in the system of $14.7~MPa$ is located where the aluminum rods mate with the pontoons.  The mount experiences the largest displacement at the bottom plate, where the maximum displacement is $0.25~mm$. See Fig.~\ref{fig:FEA1} for more details.

\begin{equation}\label{eq:safet}
    S.F = \frac{Yield\; Strength}{Maximum\; Stress} = \frac{120}{14.7} = 8.16 
\end{equation}





\subsection{Probe and Winch}
%
The central feature of our design is the sensing probe and the winch which controls the position of the probe in the $z$ axis. The probe facilitates water sample collection and provides real-time observational feedback for users, high-level controllers on the ASV (which may be important for adaptive sampling missions), and the winch controller.  The system is designed so that the probe is interchangeable though we do require that the selected probe be capable of communicating its observed depth through our tether. To be compatible with the winch, the probe assembly must be negatively buoyant but weigh less than $11~kg$ in air.  

In our experiments, a YSI EXO1 sonde~\cite{exo_sonde} (see Fig. \ref{fig:sonde}) is utilized as our probe. The sonde is connected to one end of BlueRobotic's Fathom tether~\cite{fathom_tether} by a 6-pin waterproof male-female connector. The tether provides power and a communication interface for the sonde. To reduce external drag forces on the cable, we install a cable support grip made of stainless steel onto end mated to the probe. The tether length is limited to $10m$ by the diameter of the winch spool. Future designs should consider employing either a larger spool or a thinner cable to facilitate the collection of deeper samples. 

\begin{figure*}[h!]
    \centering
    \includegraphics[width=\textwidth]{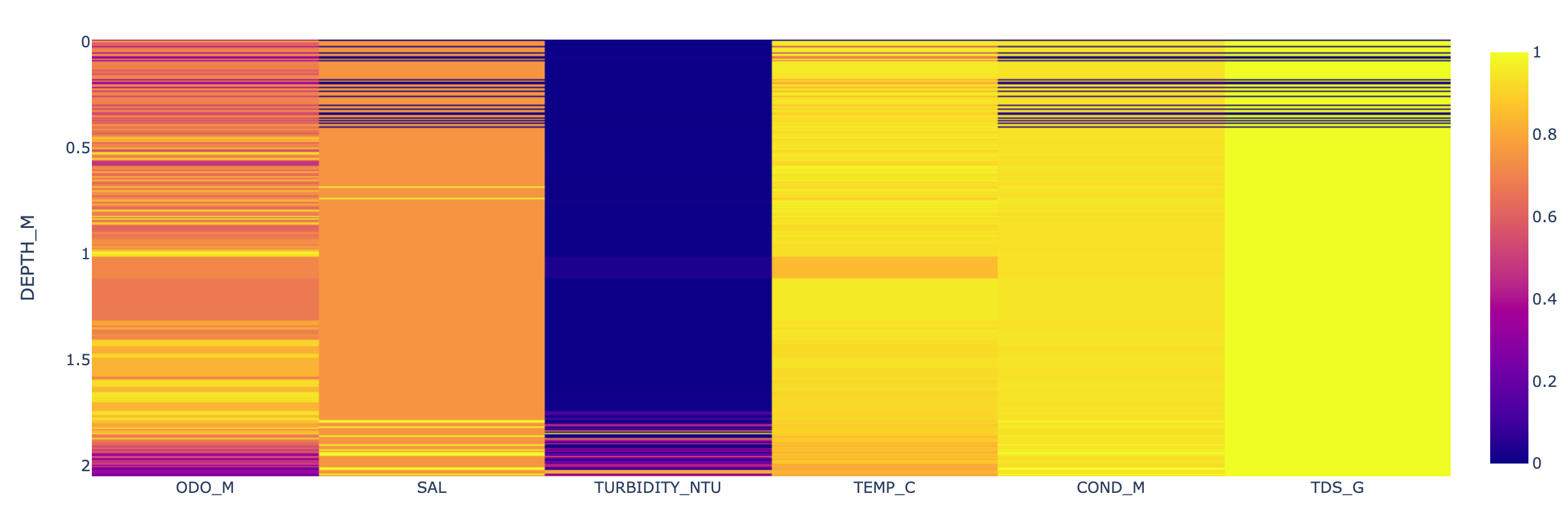}
    \caption{This figure depicts a histogram of the observed science values when normalized and compared to depth. All of the parameters we measured show reliable change over depth, emphasizing the importance of measuring throughout the water column. }
    \label{fig:depth_correlation}
\end{figure*}

The probe can be configured via the winch into two modes: \emph{underway mode} and \emph{deployed mode}. The probe is retrieved so that it sits just below the water in a vertical orientation in the \emph{underway} configuration. In the \emph{deployed} configuration, the ASV is commanded to remain stationary to reduce strain on the cable as the the winch controls the vertical position of the probe.  


We utilise the low-cost and lightweight, \emph{Trac G3 Fisherman $25$} Electric Anchor Winch (see Fig. \ref{fig:winchcam}) for probe deployment and retrieval. The winch ships with a nylon rope which we replace with $10~m$ BlueRobotics' Fathom tether~\cite{fathom_tether} to serve as both a communication channel and a physical tether.  The cable is neutrally buoyant, has $158 kg$  breaking strength, and is embedded with water-blocking fibers to seal any leaks. The tether carries four unshielded twisted pairs (UTP) of $26~AWG$ wire, which makes it compatible with several sensors.  A slip ring (see Fig.~\ref{fig:slip}) is used to transmit electrical signals from a rotating cable to the on-board computer.

Fig.~\ref{fig:winchcam} shows the configuration of the winch from our \emph{winchcam}, a ROS-enabled camera which captures a live view of the probe deployment system at $30~fps$.


\begin{table}[h]
\resizebox{.5\textwidth}{!}{%
\begin{tabular}{|l|l|l|l|}
\hline
\begin{tabular}[c]{@{}l@{}}Maximum \\ Payload (kg)\end{tabular} & \begin{tabular}[c]{@{}l@{}}Payout \\ Speed (m/min)\end{tabular} & \begin{tabular}[c]{@{}l@{}}Retrieval \\ Speed (m/min)\end{tabular} &  \begin{tabular}[c]{@{}l@{}}Operating \\ Voltage (V)\end{tabular} \\ \hline
11.340                                                          & 21.336                                                          & 19.812                                                          & 12                                                               \\ \hline
\end{tabular}%
}
\caption{APS Specifications}
\label{tab:my-table}
\end{table}

The APS system assumes the vehicle stops and maintains its station to deploy and recover the probe.  The controller considers the probe's state which is found from fusing the GPS estimate from onboard the ASV and a depth-pressure from the the probe itself. Our closed-loop controller (see Fig. \ref{fig:controller}) enables precise depth control of the sensor by deploying or retrieving the cable connecting the sensor to the ASV. Control signals are sent from the ASV's onboard computer to the an Arduino are sent through a two-channel relay to control actuation. Our system is capable of $19.812~m/min$ payout. 

\subsection{Electronics, Control, and ASV Integration}
The electronic components of the APS enable seamless communication between the probe, winch, and the central computing platform.  In our experiments, we powered both the winch and the sonde on independent, external batteries, negating the need to integrate with shipboard power. 
The winch is controlled using a relay module with two mechanical relays that allow ON-OFF-ON style control in either direction. This binary system allows power to flow from external batteries when in the \emph{on} position.  The relays are controlled by an Arduino-compatible microcontroller (MCU), which is  exposed on the ASV's internal ROS network over a serial connection.

The MCU, running its own distributed ROS node, provides functionality for both autonomous and manual control of the winch. Manual control is accomplished through open-loop service calls to raise or lower the winch in finite steps. In autonomous control mode, the system integrates real-time depth positions as reported by the probe to determine winch direction using closed-loop control. 

All interfaces utilize standard connectors. The tether to the probe provides a powered connection to the boat and supports several interfaces such as RS-232 and some Ethernet standards. The serial interface to the MCU is managed over universal serial bus (USB). These standardized connections mean that the entire system is "plug-and-play" and ROS compatible with any ASV that satisfies the payload requirements.


\section{Control Simulator}
\label{sec:simulator}
In order to develop and test our controller without deploying the boat, we built a ROS-compatible simulation environment in Unity (see Fig:\ref{fig:simulator}). 
We release this system as an open-source codebase on \href{https://github.com/edwardming789/winch_controller}{\color{blue} github}. Our simulated environment consists of a fully controllable ASV and winch equipped with a probe capable of reporting its depth. Although useful for controller evaluation, there are several limitations. 

Our simulated winch assumes a constant speed of payout/retraction. We attempt to make the probe's simulation realistic by considering gravity, buoyancy, and a drag force. Accordingly, the probe accelerates during payout until the external forces are balanced. When the winch recovers the probe, it raises to the water surface with a constant retrieval speed of 19.812$m/min$. 




\section{Experiments}
\label{sec:experiments}
\begin{figure}[h]
\begin{subfigure}[b]{0.95\columnwidth}
\centering
  \includegraphics[width=.95\columnwidth]{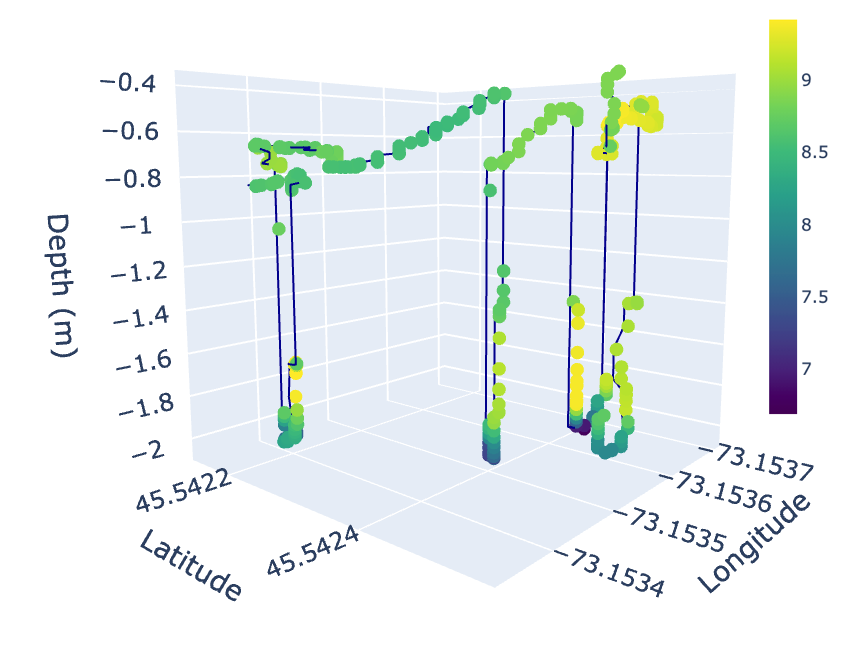}
  \caption{Dissolved Oxygen captured from our APS in four consecutive probes.  }
  \label{fig:ODO_depth}
\end{subfigure}
\begin{subfigure}[b]{0.95\columnwidth}
  \centering
  \includegraphics[width=.95\columnwidth]{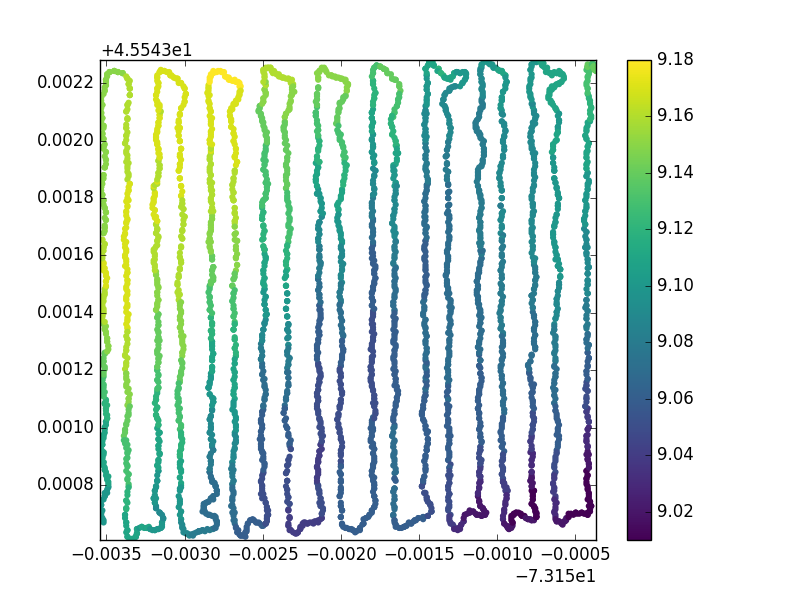}
  \caption{Dissolved Oxygen captured in the \emph{underway} mode.}  
  \label{fig:ODO_spatial}
\end{subfigure}
\label{fig:data}
\caption{Data resulting from field experiments at Lake Hertel. These plots depict one of the water quality parameters measured by our sonde probe. The data clearly varies over the surface of the lake as well as throughout the water column. }
\end{figure}

We validate our system in Lake Hertel $(45.54437, -73.15212)$ at the McGill, Gault Nature Reserve in Mont-Saint-Hilaire, Quebec, Canada. 
For the experiments in this paper, we demonstrate the APS with a 
YSI EXO multiparameter water quality sonde, equipped sensors for turbidity, temperature, salinity, PH, ORP, dissolved oxygen, conductivity, and depth. When the ASV is operating in  \emph{underway mode}, our mounting system allows continuous data collection by positioning the sensor probe just below the waterline. We  deployed the ASV to desirable locations and deployed the probe throughout the water column. See Fig.~\ref{fig:ODO_depth} for an example profile. Data from underway experiments are presented in Fig.~\ref{fig:ODO_spatial}. 

Our design was motivated by the scientific imperative to collect samples from throughout the water column from a mobile, low-cost surface vehicle. Data collected from the experiment site (see Fig.~\ref{fig:depth_correlation}) demonstrates that our system is  capable of collecting samples of phenomena which varies both spatially across the surface of the water and by depth. The samples are collected efficiently and without requiring personnel to be deployed on a boat, or even to the site given internet networking is available.  

\section{Limitations}
\label{sec:limitation}
Although the APS allows us to reach locations that are unattainable from a traditional ASV, there are some limitations. As there is no active depth control on the probe itself, we can only indirectly control the probe's position in the $z$ plane by releasing or retracting the tether. Several times during our field deployments, the probe settled on top of vegetation rather than sinking to the bottom. Without an active probe, there is little our system can do to force the probe past the vegetation and down to its target depth.

We are also limited in our localization of the probe, since we only have its depth and an estimate of the payout of the tether. In the current system, we simple assign the ASV's latitude and longitude positions to sonde readings along with the measured depth of the probe.  Future work could try to estimate the full 6DOF pose of the probe based on images or inertial measurement estimates. 

On the controls side, the current system has no behavior for realizing when the probe has become entangled. Instead, it employs a timeout function, whereby if the sensor's depth doesn't change after a configurable number of seconds of the winch operating in a direction, the controller assumes that the command must be aborted. Future versions of our system should consider a more complete version of the state and attempt to reason through possible causes of the sensor's stall (such as becoming entangled in vegetation). 

\section{Conclusion}
This paper presents a robotic system capable of collecting in situ samples from throughout the water column from an autonomous surface vehicle.  
Our solution consists of open-source mechanical, electrical, and software design to solve the problem of collecting samples at-depth. We demonstrate the efficacy of the system in lake trials, capturing data that further motivates our system. Finally, we discuss limitations and suggest future improvements. 

We believe this systems fills an important niche, enabling science-critical access to the data below the surface without requiring expensive installations or sophisticated equipment.

\section{Acknowledgements}
This work was generously supported by the NSERC Canadian Robotics Network (NCRN). We would also like to thank Chris Jing and Khaled Al Masaid for their assistance in the early development of the mechanical design. 


\bibliographystyle{IEEEtran}
\bibliography{IEEEabrv,bib}

\vspace{12pt}
\color{red}

\end{document}